\documentclass{article}

\usepackage{microtype}
\usepackage{graphicx}
\usepackage{subfigure}
\usepackage{booktabs} 
\usepackage{times}
\usepackage{epsfig}
\usepackage{amsmath}
\usepackage{amssymb}
\usepackage{bm}
\usepackage{amsthm}
\usepackage{multirow}
\usepackage{caption}
\usepackage[super]{nth}

\usepackage{amssymb}%
\usepackage{mathrsfs}%
\usepackage{nicefrac}
\usepackage{color,xcolor}
\usepackage{pifont}
\usepackage{array}
\usepackage{wrapfig}
\usepackage{mathtools}
\usepackage{colortbl}

\usepackage{tikz}
\usepackage[tikz]{bclogo}
\usepackage{pgfplots}
\pgfplotsset{width=1.0\columnwidth,compat=1.16}

\usepackage{hyperref}
\definecolor{Gray}{gray}{0.9}
\usepackage{xcolor}
\usepackage{makecell}

\usepackage[accepted]{icml2023}

\newcommand\scalemath[2]{\scalebox{#1}{\mbox{\ensuremath{\displaystyle #2}}}}

\definecolor{upurple}{RGB}{155,89,182}
\definecolor{ublue}{RGB}{52,152,219}
\definecolor{ured}{RGB}{231,76,60}
\definecolor{udark}{RGB}{77,153,77}
\definecolor{ugreen}{RGB}{46,204,113}

\newtheorem{definition}{Definition}

\hypersetup{
	colorlinks=true,
	urlcolor=magenta,
}

\icmltitlerunning{Bag of Tricks for Training Data Extraction from Language Models}

\begin{document}

\twocolumn[
\icmltitle{Bag of Tricks for Training Data Extraction from Language Models}



\icmlsetsymbol{claim}{*}

\begin{icmlauthorlist}
\vspace{-0.cm}
\icmlauthor{Weichen Yu}{claim,to1}
\icmlauthor{Tianyu Pang}{to2}
\icmlauthor{Qian Liu}{to2}
\icmlauthor{Chao Du}{to2}
\icmlauthor{Bingyi Kang}{to2}
\icmlauthor{Yan Huang}{to1}
\icmlauthor{Min Lin}{to2}
\icmlauthor{Shuicheng Yan}{to2}
\vspace{-0.cm}
\end{icmlauthorlist}

\icmlaffiliation{to1}{Institute of Automation, Chinese Academy of Sciences.}
\icmlaffiliation{to2}{Sea AI Lab}

\icmlcorrespondingauthor{Tianyu Pang}{tianyupang@sea.com}
\icmlcorrespondingauthor{Qian Liu}{liuqian@sea.com}
\icmlcorrespondingauthor{Yan Huang}{yhuang@nlpr.ia.ac.cn}

\icmlkeywords{Machine Learning, ICML}
\vskip 0.3in
]



\printAffiliationsAndNotice{$^{*}$Work done during an internship at Sea AI Lab.} 

\begin{abstract}
With the advance of language models, privacy protection is receiving more attention. Training data extraction is therefore of great importance, as it can serve as a potential tool to assess privacy leakage. However, due to the difficulty of this task, most of the existing methods are proof-of-concept and still not effective enough. In this paper, we investigate and benchmark tricks for improving training data extraction using a publicly available dataset. Because most existing extraction methods use a pipeline of generating-then-ranking, i.e., generating text candidates as potential training data and then ranking them based on specific criteria, our research focuses on the tricks for both text generation (e.g., sampling strategy) and text ranking (e.g., token-level criteria). The experimental results show that several previously overlooked tricks can be crucial to the success of training data extraction. Based on the GPT-Neo 1.3B evaluation results, our proposed tricks outperform the baseline by a large margin in most cases, providing a much stronger baseline for future research. The code is available at \href{https://github.com/weichen-yu/LM-Extraction}{https://github.com/weichen-yu/LM-Extraction}.

\end{abstract}

\vspace{-0.5cm}
\section{Introduction}
\vspace{-0.0cm}

\begin{figure}
    \centering
    \includegraphics[width=0.45\textwidth]{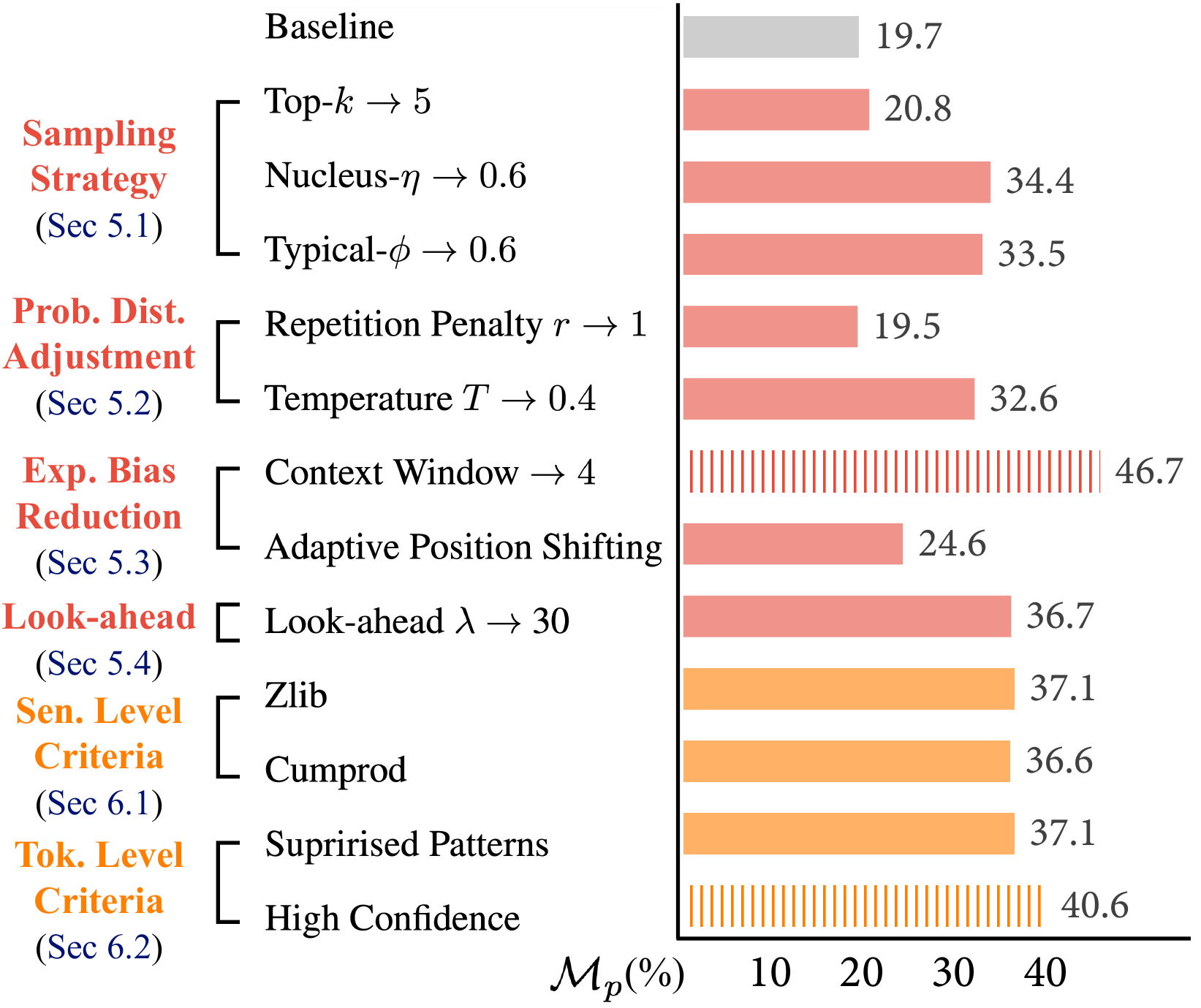}
    \vspace{-0.3cm}
    \caption{Overview for the bag of tricks explored in this work, with an evaluation of precision ($\mathcal{M}_P$). Bars in \textcolor{ured}{pink} denote the methods in the improved suffix generation, and bars in \textcolor{orange}{orange} denote the methods in the improved suffix ranking. The \emph{dashed bars} indicate the best method in each category.}
    \vspace{-0.3cm}
    \label{fig: overview}
\end{figure}

Recent advances in language models (LMs) have led to impressive performance in a variety of downstream language tasks~\citep{kenton2019bert,brown2020language}. It has been demonstrated, however, that training data can be extracted from LMs due to the memorization effects~\citep{kenton2019bert,carlini2019secret,feldman2020does,brown2020language}. These training data may contain sensitive information such as names, email addresses, phone numbers, and physical addresses, resulting in privacy leakage that hinders the widespread adoption of LMs~\citep{carlini2021extracting,carlini2022quantifying,lehman2021does}.

As privacy security has been an important issue of public concern, a crucial topic is to develop efficient methods for evaluating privacy leakage. Thus, the focus of our research is on the adversarial task of training data extraction from LMs, a relatively new area of study~\citep{carlini2021extracting,lehman2021does}. Existing extraction methods have yielded successful records, but there are instances in which these methods are even less effective than simply selecting the most popular entity based on prior score. In addition, successful data extraction requires a high generation ratio, i.e., the need to generate and rank a large number of candidates in order to identify a single successful instance. These suboptimal results suggest that, despite the viability of training data extraction and developed pioneering methods as demonstrated in previous research, this task is still relatively new with an abundance of problems to solve.

In this study, we aim to develop techniques for efficient training data extraction. We adhere to the criteria of the recent Training Data Extraction Challenge,\footnote{\href{https://github.com/google-research/lm-extraction-benchmark}{Website link} of Training Data Extraction Challenge} which employs a 1.3B parameter GPT-Neo model~\citep{gpt-neo} for \emph{targeted extraction of $1$-eidetic memorized data}. Targeted extraction refers to the scenario where a prefix of the data is provided, such as `Yu's phone number is', and the adversary attempts to recover the suffix `12345'. Accroding to \citet{carlini2021extracting}, $\kappa$-eidetic memorization is defined as the capacity of a language model to memorize a string that appears $\kappa$ times in the training material. The targeted and $1$-eidetic extraction poses a greater risk and is more challenging than non-targeted and $\kappa$-eidetic (for $\kappa>1$) settings.

Through ablation studies, we assess a variety of simple techniques in natural language processing (NLP) and empirically evaluate their impact on successful extraction rates, as in Figure~\ref{fig: overview}. Our empirical analysis reveals that the extraction performance may be sensitive to the experimental setup.
With proper settings, the results show that several previously overlooked tricks can contribute to significant improvements of training data extraction.
Based on the GPT-Neo 1.3B evaluation results, our proposed tricks outperform the baseline by a large margin in most cases, providing a much stronger baseline for future research.
Nonetheless, utilizing more than one training data extraction trick does not necessarily boost the performance, and in some cases, even shows incompatibility and results in inferior precision. These findings suggest that judicious selection and combination of the tricks are essential for optimal performance.

\vspace{-0.cm}
\section{Related Work}
\vspace{-0.0cm}
We briefly introduce training data extraction, membership inference attacks, and other memorization-based attacks.

\vspace{-0.cm}
\subsection{Training Data Extraction}
\vspace{-0.0cm}
The extraction of training data from a pretrained language model, also referred to as language model data extraction, is a method for recovering the examples used to train the model. Despite being a relatively new task, many of the underlying technologies and analysis methods, including membership inference~\citep{shokri2017membership} and leveraging network memorization for attacks~\citep{thomas2020investigating,leino2020stolen}, were introduced much earlier.

\citet{carlini2021extracting} were among the first to define the concepts of model knowledge extraction and $\kappa$-eidetic memorization, as well as to propose promising training strategies for data extraction. Both the theoretical properties of memorization and the application of model extraction in sensitive fields, such as the analysis of clinical notes, have been the focus of subsequent research in this field.

Recently, \citet{kandpal2022deduplicating} demonstrated that in language models, the efficacy of data extraction is frequently attributable to duplication in commonly used web-scraped training sets. Using nondeterminism, \citet{jagielski2022measuring} provided an explanation for forgetting memorized examples. \citet{carlini2022quantifying} analyzed three factors that affect the memorization of training data. For natural data distributions, \citet{feldman2020does} showed that label memorization is required to achieve near-optimal performance. In the application of model extraction, \citet{lehman2021does} indicated that pretrained BERT, when trained on clinical notes, poses a risk of sensitive data leakage, especially when the data exhibits a high level of repetition or `note bloat'~\citep{liu2022note}. \citet{jayaraman2022active} proposed an active extraction attack that employs canonical patterns and differential privacy to defend against pattern extraction attacks.

\vspace{-0.cm}
\subsection{Membership Inference Attacks}
\vspace{-0.0cm}
The membership inference attack (MIA)~\citep{shokri2017membership} is another adversarial task in data protection that is closely associated with training data extraction. It aims to determine whether a particular record is present in its training dataset, given black-box access to a model. MIA has been demonstrated to be effective on numerous machine learning tasks, including classification~\citep{sablayrolles2019white,choquette2021label,rezaei2021difficulty} and generation~\citep{hayes2019logan,hilprecht2019monte}.


The methods utilized by MIA fall into two categories: classifier-based methods and metric-based methods~\citep{hu2022membership}. Classifier-based methods involve training a binary classifier to recognize the complex relationship between members and non-members, with shadow training being a commonly used technique~\citep{shokri2017membership,he2020segmentations,wang2021against}. Metric-based methods, on the other hand, make membership inferences by first calculating metrics on the model prediction vectors~\citep{yeom2018privacy,salem2018ml,sablayrolles2019white,song2021systematic,choquette2021label}. Several defense methods based on differential privacy~\citep{leino2020stolen,naseri2020toward,choquette2021label}, data pruning~\citep{wang2021against}, data augmentation~\citep{kaya2021does} and causal inference~\citep{tople2020alleviating} have been proposed to mitigate this vulnerability.

\vspace{-0.cm}
\subsection{Other Memorization-Based Attacks}
\vspace{-0.0cm}
It has been discovered that large pretrained models are susceptible to memorizing information from the training data, which can lead to a variety of attacks. In addition to training data extraction and membership inference attacks, there are other memorization-based attacks that target these models.

Model extraction attacks and the corresponding protection methods~\citep{tramer2016stealing,juuti2019prada,gong2020model,wu2022model} focus on the issue of duplicating the functionality of a given model. In this type of attacks, the adversary attempts to build a second model with a similar predictive performance to the original black-box model.

The objective of attribute inference attacks is to extract specific personal attributes such as locations, occupations, and interests from the model~\citep{fredrikson2015model,gong2016you,ganju2018property,parisot2021property}. The objective of property inference attacks is to extract properties of the training data that the model producer may not have intended to share, such as the environment in which the data was generated or the proportion of the data that belongs to a particular class. The primary distinction between training data extraction attacks and attribute/property inference attacks is that attribute/property inference attacks do not require prior knowledge of the attributes or properties to be extracted, whereas training data extraction attacks require the generated information to be identical to the training data at the sentence level, which is more difficult and dangerous.

\vspace{-0.cm}
\section{Preliminary} \label{sect: definition}
\vspace{-0.0cm}

We recap the basic setups employed in our study. These setups mainly follow the guidelines of the Training Data Extraction Challenge. We then define the threat model and evaluation metrics.

\vspace{-0.cm}
\subsection{Basic Setups}
\vspace{-0.0cm}
\textbf{Dataset.} The dataset used in this study is a subset of 20,000 examples from the Pile's training dataset~\citep{gao2020pile}. Each example consists of a 50-token prefix and a 50-token suffix. The attacker's task is to predict the suffix given the prefix. All the 100-token long sentences in this dataset appear only once in the training set. For the purposes of this study, we divide the dataset into a training set of 19,000 samples and a testing set of 1,000 samples.

\textbf{Language model.} We employ the GPT-Neo 1.3B model implemented on HuggingFace Transformers~\citep{wolf2020transformers}, which is a transformer model designed using EleutherAI's replication of the GPT-3 architecture~\citep{brown2020language}, and trained on the Pile dataset. GPT-Neo is an autoregressive language model $f_{\theta}$ parameterized by $\theta$, which generates a sequence of tokens $x_{0},x_{1},\cdots,x_{N}$ via the chain rule
\begin{equation}
f_{\theta}\left(x_{0},x_{1},\cdots,x_{N}\right)=\prod_{n=0}^{N}f_{\theta}\left(x_{n}|x_{[0,n-1]}\right)\textrm{,}
\end{equation}
where $x_{[0,n-1]}=x_{<n}=\{x_{0},\cdots,x_{n-1}\}$ for notation compactness. At the sentence level, given a prefix $p$, we denote the probability of generating a certain suffix $s$ conditional on the prefix $p$ as $f_{\theta}(s|p)$.


\vspace{-0.cm}
\subsection{Threat Model}
\vspace{-0.0cm}
In this study, we focus on the threat model of targeted extraction of $\kappa$-eidetic memorized data, where we choose $\kappa=1$. According to the model knowledge extraction defined in \citep{carlini2021extracting}, we assume the language model generates a suffix $s$ by the most-likely criterion. Then we can write a formal definition of targeted extraction as
\begin{definition}
(Targeted extraction) Given a prefix $p$ contained in the training data and a pretrained language model $f_{\theta}$. 
Targeted extraction is to generate the suffix by $s=\textrm{argmax}_{s'} f_{\theta}(s'|p)$.
\end{definition}

As to $\kappa$-eidetic memorized data, we follow the definition in \citet{carlini2021extracting} that the sentence $[p,s]$ appears in at most $\kappa$ examples in the training data. In practice, the length of the generated sentence is typically fixed using truncating and concatenation techniques applied to the training dataset. If a generated sentence is shorter than the specified length, padding tokens are used to bring it up to the required length. In this study, the generated sentence length is 100.

\vspace{-0.cm}
\subsection{Evaluation Metrics}
\vspace{-0.cm}
Non-targeted extraction has been evaluated using the number of memorized examples in previous studies~\citep{carlini2021extracting}. To evaluate more comprehensively, we use three metrics to evaluate performance in this targeted data extraction task, including precision $\mathcal{M}_{\textrm{P}}$, recall $\mathcal{M}_{\textrm{R}}$ and Hamming distance $\mathcal{M}_{\textrm{H}}$.


\textbf{Precision $\mathcal{M}_{\textrm{P}}$.} 
The proportion of correctly generated suffixes over the total number of given prefixes is referred to as precision $\mathcal{M}_{\textrm{P}}$. Notice that for a correct generation, the suffix and ground truth must be identical in both sentence length and generated tokens.



\textbf{Recall $\mathcal{M}_{\textrm{R}}$.} The proportion of correctly generated suffixes over the total number of generated suffixes is indicated by recall $\mathcal{M}_{\textrm{R}}$. The metric used in the Training Data Extraction Challenge is denoted by $\boldsymbol{e}_{j}$, which is defined as the number of correctly recovered suffixes when the number of incorrectly generated suffixes reaches a threshold of $j$. $\boldsymbol{e}_{j}$ can assess the effectiveness of the attack. We define $\mathcal{M}_{\textrm{R}}=\frac{\boldsymbol{e}_{j}}{\boldsymbol{e}_{j}+j}$, we will use $\mathcal{M}_{\textrm{R}}$ instead of $e_j$ in the following paragraphs. In our experiments, the value of $j$ is chosen to be proportional to the size of the test set, and it is set to 100 with a test set of 1,000 prefixes.


\textbf{Hamming distance $\mathcal{M}_{\textrm{H}}$.} The Hamming distance denotes the difference between two equal-length strings, calculated as the number of positions where the corresponding symbols differ. We can quantitatively evaluate the similarity between the generated suffixes and the ground truth using the average Hamming distance, providing a token-level evaluation of the extraction methods' performance. $\mathcal{M}_{\textrm{H}} = \frac{1}{N}\sum_{n} x_n\oplus gt_n$, where $a\oplus b=1 \;\textrm{if}\; a=b, \textrm{else}\; 0$. $N$ is the number of tokens in a generated sentence, $x_n$ is the generated token, and $gt_n$ is the corresponding ground truth token.

We would like to point out that the commonly used metrics that measure the repetition or quality of generated sentences, e.g., the Zipfian coefficient and REP~\citep{welleck2019neural}, may not be aligned with the goal of training data extraction. Thus, we propose the metrics listed above, which are specifically designed to assess the efficacy of training data extraction methods. These metrics, $\mathcal{M}_{\textrm{P}}$, $\mathcal{M}_{\textrm{R}}$, and $\mathcal{M}_{\textrm{H}}$, are closely linked with the success rate of the extracted data rather than the quality of generated language.


\vspace{-0.cm}
\section{Pipeline Overview}
\vspace{-0.cm}
A basic training data extraction pipeline can be divided into two stages. The first stage is \emph{suffix generation}, which means coming up with a set of suffixes based on a prefix. The autoregressive language model $f_{\theta}$ computes the probability of each token in a vocabulary list and generates the next token by sampling from the probability distribution. A basic strategy employed to control the number of generations is limiting to the top-$k$ tokens with $k=10$, which means the LMs only sample from the tokens with top-$k$ probability. The second stage is \emph{suffix ranking}, which entails estimating the likelihood of these suffixes and retaining only those with a high probability. Typically this is accomplished through the use of membership inference attacks, determined based on the perplexity metric~\citep{carlini2019secret,carlini2021extracting} as
\begin{equation}
    \mathcal{P} = \exp\left(-\frac{1}{N}\sum_{n=0}^{N} \log f_{\theta}(x_n|x_{[0:n-1]})\right)\textrm{,}
\end{equation}
where $N$ is the number of tokens in a generated sentence. Here, padding tokens are not included in the calculation. We explore and evaluate various tricks for both stages. The results for suffix generation are presented in Section~\ref{sect: tricks}, and the results for suffix ranking are presented in Section~\ref{sect: ranking}.




\section{Improved Suffix Generation} \label{sect: tricks}



\begin{figure}[t]
    \centering
    \includegraphics[width=.98\linewidth]{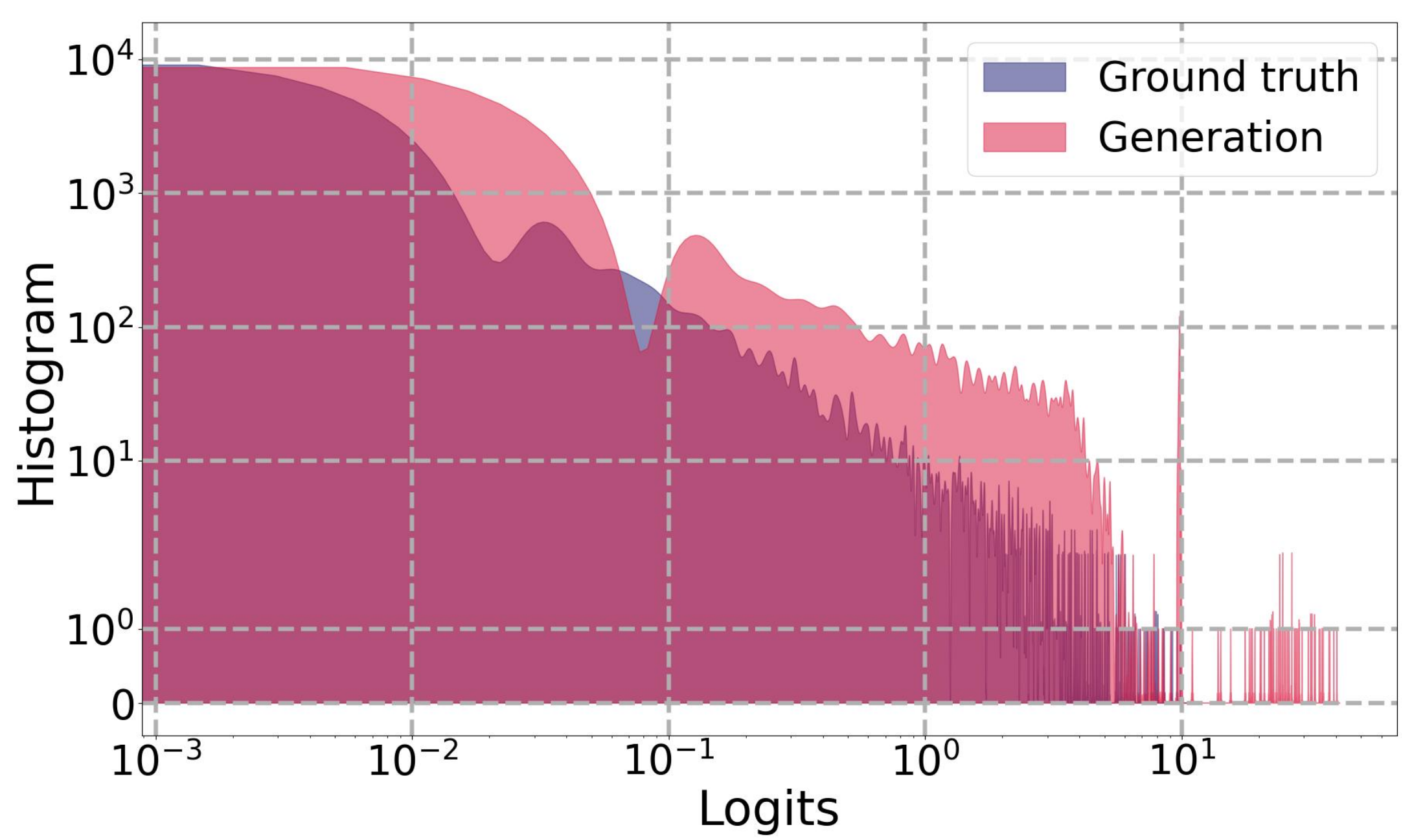}
    \vspace{-0.3cm}
    \caption{Histogram of token logits. The histogram depicts the distribution of logit values obtained from 1,000 suffixes, each containing 50 tokens. A spline interpolation technique is employed to smooth the histogram, with the original histogram included in the appendix for reference.}
    \label{fig:token_hist_spline}
\end{figure}

To improve suffix generation, we analyze the logits distributions of both the ground-truth and generated tokens. As shown in Figure~\ref{fig:token_hist_spline}, there is a significant difference between the two distributions. To address this, we examine the effects of a variety of NLP techniques, such as changes to sampling strategies and probability distributions, as described below.

\subsection{Sampling Strategy} \label{sect:sampling}


{The most popular decoding objective, especially for conditional generation in NLP tasks, is maximization-based decoding. Based on the provided prefix, it searches for the suffix with the highest likelihood. It is also suited for our training data extraction task since the models are directly maximized for the likelihood of the training data. However, finding the theoretically argmax sequence from models is intractable~\citep{holtzman2019curious}. The common practice is using beam search~\citep{freitag-al-onaizan-2017-beam} instead.

As an approximate solution, beam search only keeps a predetermined number of best partial solutions. 
Since it always takes the top ones at each step, it is often criticized for its lack of diversity~\citep{holtzman2019curious}.
As shown in Table~\ref{tab:beams}, although the required memory is increasing, the performance gain from a larger beam width diminishes quickly when the beam width exceeds two.
Due to the randomness in generation, sampling methods always introduce more diversity than beam search.
Therefore, we mainly study sampling strategies below, including top-$k$ sampling, nucleus sampling and typical sampling.
}

\begin{table}[t]
\centering
\small
\caption{Results of $\mathcal{M}_{\textrm{P}}$, $\mathcal{M}_{\textrm{R}}$, and $\mathcal{M}_{\textrm{H}}$ under different numbers of beams in beam search. All results are reported on a single trial.}
\vspace{0.1in}
\begin{tabular}{crccc}
\toprule
\textbf{Beam} & \textbf{Memory} & \makecell[c]{$\mathcal{M}_{\textrm{P}}$(\%) $\uparrow$} & \makecell[c]{$\mathcal{M}_{\textrm{R}}$(\%) $\uparrow$} & \makecell[c]{$\mathcal{M}_{\textrm{H}}$ $\downarrow$} \\ \midrule
7 & 21531M & 51.1               & 76.3        & 15.501  \\
6 & 21379M & 51.0               & 76.4       & 15.575  \\
5 & 20731M & 51.3               & 76.3         & 15.499  \\
4 & 17051M & 50.7               & 76.3        & 15.763  \\
3 & 14333M & 50.7               & 76.2        & 15.764  \\
2 & 12023M & 50.6               & 76.2        & 15.931  \\
1 & 8783M  & 37.0               & 76.5        & 20.245 \\
    \bottomrule
\end{tabular}
\label{tab:beams}
\vspace{-.1cm}
\end{table}

\begin{figure}[bt]
    \centering
        \begin{tikzpicture}[scale=0.75]
        \begin{axis}[
        width=.3\textwidth,
        height=.25\textwidth,
        ymin=36, ymax=43,
        axis y line*=left,
        xmin=-4, xmax=49,
        xtick={0,10,...,40},
        ytick={37,38,39,...,42},
        y tick style={opacity=0},
        xlabel={(a) Top-k},
        xlabel style={align=center,yshift=0.5em},
        grid style=dashed,
        ylabel={Precision (\%)},
        xlabel style={align=center},
        xlabel style={align=center},
        ylabel style={yshift=-0.5em},
        ymajorgrids=true,
        xmajorgrids=true,
        tick align=inside]

        \addplot[
            ured!60,mark=square*,mark size=2pt,line width=1.5pt,mark options={fill=white,draw=orange,line width=1pt}
            ]
            coordinates {
              (0, 38.8)
              (5, 37.0)
              (10, 36.9)
              (15, 36.8)
              (20, 36.8)
              (25, 36.6)
              (30, 36.6)
              (35, 36.5)
              (40, 36.5)
              (45, 36.5)
            };
            \label{plot_two}
            
        \end{axis}

        \begin{axis}[
          width=.3\textwidth,
          height=.25\textwidth,
          axis y line*=right,
          axis x line=none,
          xmin=-4, xmax=49,
          ymin=19.4, ymax=20.8,
          ytick={19.6,19.8,20.0,20.2,20.4,20.6},
          yticklabels={$19.6$,$19.8$,$20.0$,$20.2$,$20.4$,$20.6$},
          tick align=inside,
          y tick style={opacity=0},
          legend style={at={(1.2,1.15)},anchor=south},
          legend columns=3,
          legend cell align={left},
          clip=false
        ]
        \addlegendimage{/pgfplots/refstyle=plot_one}\addlegendentry{$\mathcal{M}_P$}
        \addplot[upurple!80,mark=triangle*,,mark size=3pt,line width=1.5pt,mark options={fill=white,draw=upurple,line width=1pt}]
            coordinates {
              (0, 19.643)
              (5, 19.614)
              (10, 20.429)
              (15, 20.368)
              (20, 20.355)
              (25, 20.47)
              (30, 20.491)
              (35, 20.506)
              (40, 20.564)
              (45, 20.555)
            };
        \addlegendentry{$\mathcal{M}_H$}
        \end{axis}
        \vspace{2mm}

        \begin{axis}[
        at={(15.5em,0)},
        width=.3\textwidth,
        height=.25\textwidth ,
        ymin=38, ymax=54,
        axis y line*=left,
        xlabel style={yshift=0ex},
        xmin=-0.1, xmax=1.1,
        xtick={0.0,0.2,...,1.0},
        ytick={40,42,...,52},
        y tick style={opacity=0},
        xlabel={(b) Nucleus-$\eta$},
        xlabel style={align=center,yshift=0.5em},
        grid style=dashed,
        ylabel={Precision (\%)},
        xlabel style={align=center},
        xlabel style={align=center},
        ylabel style={yshift=-0.5em},
        ymajorgrids=true,
        xmajorgrids=true,
        tick align=inside]

        \addplot[
            ured!60,mark=square*,mark size=2pt,line width=1.5pt,mark options={fill=white,draw=orange,line width=1pt}
            ]
            coordinates {
              (0.99, 39.1)
              (0.95, 42.3)
              (0.90, 44.1)
              (0.8, 47.5)
              (0.7, 48.1)
              (0.6, 48.5)
              (0.5, 47.3)
              (0.4, 46.9)
              (0.3, 46.2)
              (0.2, 46.2)
              (0.1, 46.2)
            };
            
        \end{axis}

        \begin{axis}[
          at={(15.5em,0)},
          width=.3\textwidth,
          height=.25\textwidth,
          axis y line*=right,
          axis x line=none,
          xmin=-0.1, xmax=1.1,
          ymin=16.0, ymax=20.0,
          ytick={16.5,17.0,...,19.5},
          tick align=inside,
          yticklabel style={/pgf/number format/precision=1,/pgf/number format/fixed zerofill},
          y tick style={opacity=0},
          legend style={at={(0.5,1.15)},anchor=south},
          legend columns=3,
          legend cell align={left},
          clip=false
        ]
        \addplot[upurple!80,mark=triangle*,,mark size=3pt,line width=1.5pt,mark options={fill=white,draw=upurple,line width=1pt}]
            coordinates {
              (0.99, 19.704)
              (0.95, 18.742)
              (0.9, 17.851)
              (0.8, 16.986)
              (0.7, 16.46)
              (0.6, 16.442)
              (0.5, 16.882)
              (0.4, 16.977)
              (0.3, 17.352)
              (0.2, 17.389)
              (0.1, 17.392)
            };
        \end{axis}

        \end{tikzpicture}

        \begin{tikzpicture}[scale=0.75]
        \begin{axis}[
        width=.3\textwidth,
        height=.25\textwidth,
        ymin=34, ymax=55,
        axis y line*=left,
        xmin=-0.1, xmax=1.1,
        xtick={0.0,0.2,...,1.0},
        ytick={37,40,...,52},
        y tick style={opacity=0},
        xlabel={(c) Typical-$\phi$},
        xlabel style={align=center,yshift=0},
        grid style=dashed,
        ylabel={Precision (\%)},
        ylabel style={yshift=-0.5em},
        ymajorgrids=true,
        xmajorgrids=true,
        tick align=inside]

        \addplot[
            ured!60,mark=square*,mark size=2pt,line width=1.5pt,mark options={fill=white,draw=orange,line width=1pt}
            ]
            coordinates {
              (0.1, 37.4)
              (0.2, 38.8)
              (0.3, 43.0)
              (0.4, 46.9)
              (0.5, 48.2)
              (0.6, 48.7)
              (0.7, 48.1)
              (0.8, 47.5)
              (0.9, 44.1)
              (0.95, 42.3)
              (0.99, 39.1)
            };
            \label{plot_2}
            
        \end{axis}

        \begin{axis}[
          width=.3\textwidth,
          height=.25\textwidth,
          axis y line*=right,
          axis x line=none,
          xmin=-0.1, xmax=1.1,
          ymin=15.5, ymax=22.5,
          ytick={16.5,17.5,...,21.5},
          tick align=inside,
          y tick style={opacity=0},
          yticklabel style={/pgf/number format/precision=1,/pgf/number format/fixed zerofill},
        ]
        \addplot[upurple!80,mark=triangle*,,mark size=3pt,line width=1.5pt,mark options={fill=white,draw=upurple,line width=1pt}]
            coordinates {            
              (0.1, 21.336)
              (0.2, 20.416)
              (0.3, 18.655)
              (0.4, 17.062)
              (0.5, 16.636)
              (0.6, 16.499)
              (0.7, 16.428)
              (0.8, 17.023)
              (0.9, 17.854)
              (0.95, 18.746)
              (0.99, 19.701)
            };
        \end{axis}
        \vspace{2mm}

        \begin{axis}[
        at={(15.5em,0)},
        width=.3\textwidth,
        height=.25\textwidth ,
        ymin=-10, ymax=60,
        axis y line*=left,
        xlabel style={yshift=0ex},
        xmin=0, xmax=20,
        xtick={0.1, 1.0, 10.0},
        xticklabels={$10^{-1}$, $10^0$, $10^1$},
        ytick={0,10,...,50},
        y tick style={opacity=0},
        xlabel={(d) Temperature},
        xlabel style={align=center,yshift=0.5em},
        grid style=dashed,
        ylabel={Precision (\%)},
        xlabel style={align=center},
        xlabel style={align=center},
        ylabel style={yshift=-0.5em},
        ymajorgrids=true,
        xmajorgrids=true,
        xmode=log,
        log ticks with fixed point,
        tick align=inside]

        \addplot[
            ured!60,mark=square*,mark size=2pt,line width=1.5pt,mark options={fill=white,draw=orange,line width=1pt}
            ]
            coordinates {
              (0.1, 47)
              (0.2, 48.2)
              (0.3, 48.9)
              (0.4, 49.4)
              (0.5, 48.4)
              (0.6, 47.6)
              (0.7, 45.9)
              (0.8, 43.6)
              (0.9, 40.1)
              (1.0, 37.0)
              (1.1, 33.8)
              (2, 6.5)
              (10, 0.0)
            };
            
        \end{axis}

        \begin{axis}[
          at={(15.5em,0)},
          width=.3\textwidth,
          height=.25\textwidth,
          axis y line*=right,
          axis x line=none,
          xmin=0, xmax=20,
          ymin=2, ymax=64,
          ytick={10,18,...,56},
          tick align=inside,
          yticklabel style={/pgf/number format/precision=1,/pgf/number format/fixed zerofill},
          y tick style={opacity=0},
          legend style={at={(0.5,1.15)},anchor=south},
          legend columns=3,
          legend cell align={left},
          xmode=log,
          log ticks with fixed point,
          clip=false
        ]
        \addplot[upurple!80,mark=triangle*,,mark size=3pt,line width=1.5pt,mark options={fill=white,draw=upurple,line width=1pt}]
            coordinates {
              (0.1, 16.899)
              (0.2, 16.508)
              (0.3, 16.341)
              (0.4, 16.213)
              (0.5, 16.437)
              (0.6, 16.646)
              (0.7, 17.075)
              (0.8, 17.979)
              (0.9, 19.018)
              (1.0, 20.245)
              (1.1, 21.469)
              (2, 34.693)
              (10, 48.984)
            };
        \end{axis}

        \end{tikzpicture}
    \vspace{-0.3cm}
    \caption{Experimental results under different values of top-$k$, nucleus-$\eta$, typical-$\phi$ and temperature $T$. All results are reported on 5 trials. The y-axis left denotes precision (\%)$\left(\uparrow\right)$, and right denotes Hamming distance $\left(\downarrow\right)$. }
    \label{fig:topketc}
\end{figure}
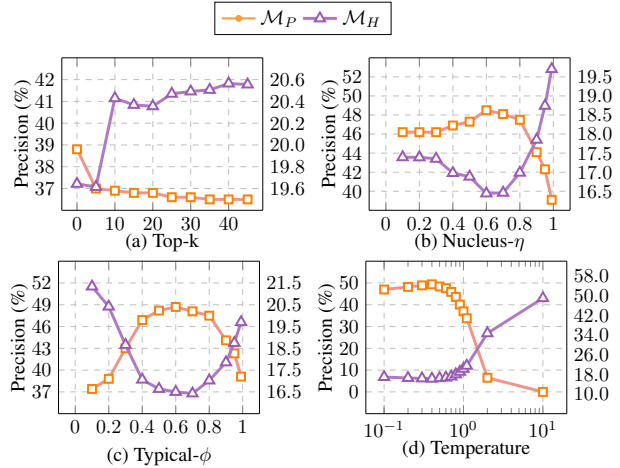

\textbf{Top-$k$} denotes the truncated sampling method with the truncation set defined as the top-$k$ highest-probability tokens~\citep{fan2018hierarchical,holtzman2018learning,radford2019language}. Figure~\ref{fig:topketc}(a) demonstrates that a larger $k$ may increase the diversity of the generated sentences but also decrease the precision and hamming distance. The result is consistent with that observed in other NLP tasks~\citet{meister2022locally}. 

\textbf{Nucleus-$\eta$}~\citep{holtzman2019curious} truncates the vocabulary based on the summed probability. Namely, the smallest set of the most likely tokens with total probabilities equal to or greater than $\eta$ are selected.
\citet{meister2022locally} demonstrate that lower values of $\eta$ are more conducive to story generation. In contrast, the extraction task is optimal at $\eta\approx 0.6$, which yields a 31\% improvement in extraction precision over the baseline. Larger or smaller $\eta$ both shows decreased performance (Figure~\ref{fig:topketc}(b)). 


\textbf{Typical-$\phi$}~\citep{meister2022locally} recommends selecting a token with information content similar to the expected information content. Namely, it guarantees the probability mass from the original distribution that will be taken into account in typical decoding. This sampling strategy can improve sentence consistency while reducing degenerate repetitions. The typical-$\phi$ strategy is equivalent to a subset optimization problem with an entropy rate constraint. Typical-$\phi$ exhibits a non-monotonic tendency, which is similar to its effect on abstract summary and story generation tasks.

\subsection{Probability Distribution Adjustment}


{Besides the sampling strategies that truncate the sampling distribution, we introduce in this section another strategy that directly adjusts the probability distribution $f_{\theta}\left(x_{n}|x_{<n}\right)$. Below we introduce two tricks, adjusting the temperature and repetition penalty, to refine the distribution. These tricks bring nearly no additional computation cost but can still lead to a performance improvement.}

\textbf{Temperature $T$}~\citep{DBLP:journals/corr/HintonVD15} is a technique of local renormalization with an annealing term. A higher temperature $T > 1$ results in decreased confidence in the language model's predictions but may also increase the diversity of the generated suffixes. The study conducted by \citet{carlini2021extracting} found that gradually decreasing the temperature throughout the generation process can be beneficial. The effect of the temperature is presented in Table~\ref{tab:temp}. It is important to note that as the temperature is increased, the number of generated suffixes required to include the ground truth also increases, causing the efficiency to degrade. It is important to find a balance between diversity and efficiency.

\textbf{Repetition penalty} is constructed on the conditional language model~\citep{keskar2019ctrl}. 
A repetition penalty is introduced by locally modifying the generation probability of each token based on whether it is a repetition of the previous token. The logit of the repeated token is divided by a value $r$ before entering the softmax layer. Setting $r > 1$ penalizes repetition while $r < 1$ encourages it. Our results in Table~\ref{tab:rp} show that repetition penalty has mostly negative effects on the task of training data extraction.

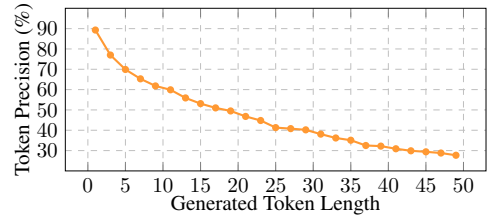
\begin{figure}[tb]
\vspace{-0.05cm}
    \centering
        \begin{tikzpicture}[scale=0.8]
        \begin{axis}[
        width=.5\textwidth,
        height=.25\textwidth,
        ymin=20, ymax=100,
        xlabel style={yshift=0ex,},
        xmin=-3, xmax=53,
        xtick={0,5,...,50},
        ytick={30,40,50,...,90},
        y tick style={opacity=0},
        xlabel={Generated Token Length},
        xlabel style={align=center,yshift=0.5em},
        grid style=dashed,
        ylabel={Token Precision (\%)},
        xlabel style={align=center},
        xlabel style={align=center},
        ylabel style={yshift=-0.5em},
        ymajorgrids=true,
        xmajorgrids=true,
        tick align=inside]
        \addplot[
            orange!80,mark=*,mark size=1.2pt,line width=1pt]
            coordinates {
             (1, 89.3)
             (3, 77.0)
             (5, 69.9)
             (7, 65.3)
             (9, 61.8)
             (11, 59.9)
             (13, 55.9)
             (15, 53.1)
             (17, 51.0)
             (19, 49.5)
             (21, 46.8)
             (23, 44.8)
             (25, 41.3)
             (27, 40.8)
             (29, 40.2)
             (31, 38.1)
             (33, 36.2)
             (35, 35.1)
             (37, 32.5)
             (39, 32.2)
             (41, 30.9)
             (43, 29.9)
             (45, 29.4)
             (47, 28.8)
             (49, 27.7)
            };
            \label{plot_one}
        \end{axis}

        \end{tikzpicture}
    \vspace{-0.3cm}
    \caption{Generated token length w.r.t. token precision (\%) for the $n$-th generated token. The generated suffix length is 50.}
    \label{fig:token preci len}
\end{figure}

\subsection{Exposure Bias Reduction}

For efficient vectorization, it is common practice to pack multiple sentences into a fixed-length sequence when training language models. As an example, consider the sentence `The phone number of Yu is 12345' may be truncated or prefixed with another sentence in the training set, such as `number of Yu is 12345' or `Yu's address is at XXX. The phone number of Yu is 12345'. The prefix in the training set, as in Table~\ref{tab:fail case}, is not always a complete sentence. To better mimic the training settings, we propose to adjust the context window size and adjust the position shifting.


\subsubsection{Dynamic Context Window}
The length of the training window may differ from the length of the extraction window. As a result, we propose adjusting the context window size, i.e. the number of previously generated tokens, as shown in Eq.~(\ref{eq:winlen}). Furthermore, we encourage the results of different context window sizes to collaborate in determining the next generated token as
\begin{equation}
    \begin{split}
       \!\!\!\! &f_{\theta}\!\left(x_{n};\mathcal{W}\right) \\
       \!\!\!\! &=\!h_{\mathcal{W}}\!\left(f_{\theta}\!\left(x_{n}|x_{[n-w_1,n-1]}\right),...,f_{\theta}\!\left(x_{n}|x_{[n-w_m,n-1]}\right)\right)\!\textrm{,} \!\!
    \end{split}
    \label{eq:winlen}
\end{equation}
where $h_{\mathcal{W}}$ denotes the ensemble method, $\mathcal{W}$ denotes the ensemble hyperparameter, including the number of different context window sizes $m$ and each window size $w_i$. We use $m=4$ and $w_i \in \{n,n-1,n-2,n-3\}$ in our methods. Carefully chosen hyperparameters $m$ and $w_i$ may improve the performance even more.

\begin{figure}[t]
    \centering
    \includegraphics[width=\linewidth]{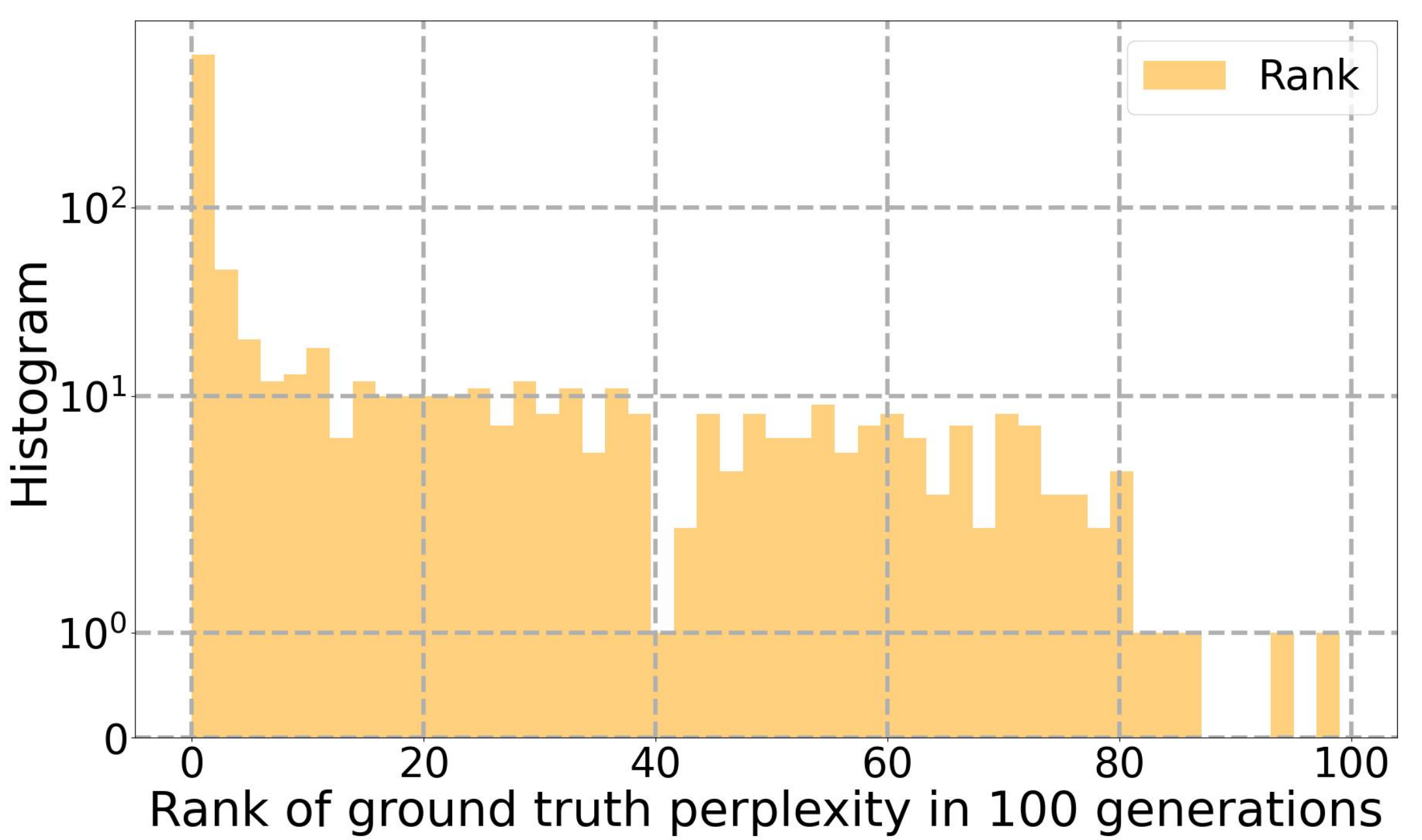}
    \vspace{-0.6cm}
    \caption{Histogram of the rank of the ground truth perplexity. The x-axis represents the rank of the ground truth perplexity within a list of 100 suffix perplexities. 
    }
    \vspace{-0.cm}
    \label{fig:sen_hist_original}
\end{figure}

We present two implementation options. The first option, as specified in Eq.~(\ref{eq:ws}), entails computing the probabilities generated by utilizing various lengths of previously generated tokens $x_{[n-w_i,n-1]}$, and then producing the final probabilities via a weighted average sum of these probabilities, as
\begin{equation} \label{eq:ws}
    f_{\theta}\left(x_{n};\mathcal{W}_w\right) = \frac{\sum_{i=1}^{m}{\epsilon_i  f_{\theta}\left(x_{n}|x_{[n-w_i,n-1]}\right)}}{\sum_{i=1}^{m}{\epsilon_i}}\textrm{,}
\end{equation} 
where $\mathcal{W}_w$ denotes the hyperparameters in the solution, comprising of $m$, $w_i$ and $\epsilon_i$. $\epsilon_i$ denotes the weighting coefficient of each probability. The second option as specified in Eq.~(\ref{eq:vote}), is based on a voting mechanism, in which each model trained with a distinct context window length casts its vote for the tokens it is most confident in, formulated as
\begin{equation} \label{eq:vote}
        f_{\theta}(x_{n};\mathcal{W}_v) = \frac{1}{m}\sum_{i=1}^{m}\mathcal{V}(f_{\theta}(x_{n}|x_{[n-w_i,n-1]})\textrm{;}
\end{equation}
\begin{equation}
\!\!\! \mathcal{V}(f_{\theta}(x_{n})) = 
        \begin{cases}
            \rho-\mathcal{R}(f_{\theta}(x_{n}))\textrm{,} &\textrm{if} \; \mathcal{R}(f_{\theta}(x_{n})) \leq \rho \textrm{;} \\
            0\textrm{,} &\textrm{otherwise} \textrm{,}
        \end{cases}
\end{equation}
where $\mathcal{V}(\cdot)$ denotes the voting function, $\mathcal{R}(\cdot)$ denotes the rank function, and it votes for the tokens that it has confidence in. $\mathcal{W}_v$ denotes the hyperparameters in the solution, comprising of $w_i$, $m$ and $\rho$.

It is stated in~\citet{carlini2022quantifying} that the proportion of extractable sequences increases log-linearly with the number of context tokens. We observed a similar phenomenon in our experiments, where the generation accuracy of a token decreases as the prefix becomes shorter. However, we discovered that combining multiple context window lengths significantly improves accuracy. The probabilities produced by different window lengths can be combined to significantly improve extraction accuracy. Our implementation of $\mathcal{W}_w$ employs the weighting coefficient $\epsilon_i=0.9^i$, and $\mathcal{W}_v$ assigns $[5,4,3,2,1]$ points to its top-$5$ confident tokens, $\rho=5$. The results presented in Table~\ref{tab: window_len} show that ensemble methods can significantly improve on the baseline approach, achieving improvements of 143\% and 139\%, respectively, and we discovered that the weighted average strategy performs slightly better than the voting strategy. One common failure mode observed is that when a wrong token is generated, it causes subsequent tokens to also be wrong (exemplars shown in Table \ref{tab:fail case}). The window size ensemble introduced here can help reduce this problem.

\begin{table}[t]
\centering
\caption{Results of $\mathcal{M}_{\textrm{P}}$, $\mathcal{M}_{\textrm{R}}$, and $\mathcal{M}_{\textrm{H}}$ under different temperature. Temperature = 1 is the baseline. All results are reported on 5 trials.}
\small
\vspace{0.1in}
\begin{tabular}{cccc}
\toprule
\textbf{Temperature} & $\mathcal{M}_{\textrm{P}}$ (\%)$\left(\uparrow\right)$    & $\mathcal{M}_{\textrm{R}}$ (\%)$\left(\uparrow\right)$ & $\mathcal{M}_{\textrm{H}}$   $\left(\downarrow\right)$     \\ \midrule
Varying     & 48.0 & 76.3 & 19.614 \\ \midrule
0.3         & 48.9 & 76.4 & 16.341 \\
1           & 37.0  & 76.5 & 20.245  \\
    \bottomrule
\end{tabular} \label{tab:temp}
\end{table}

\begin{table}[t]
\centering
\caption{Results of $\mathcal{M}_{\textrm{P}}$, $\mathcal{M}_{\textrm{R}}$, and $\mathcal{M}_{\textrm{H}}$ under different repetition penalty. Repetition penalty $r=1$ is the baseline. All results are reported on 5 trials.}
\small
\vspace{0.1in}
\begin{tabular}{cccc}
\toprule
\textbf{Repetition penalty}    & $\mathcal{M}_{\textrm{P}}$  (\%)$\left(\uparrow\right)$ & $\mathcal{M}_{\textrm{R}}$  (\%)$\left(\uparrow\right)$& $\mathcal{M}_{\textrm{H}}$ $\left(\downarrow\right)$\\ \midrule
0.9         & 19.8              & 66.4        & 27.927  \\
1           & 37.0              & 76.5       & 19.614  \\
1.1         & 37.3              & 76.5        & 20.181  \\
1.2         & 37.1              & 76.5        & 20.323  \\
1.3         & 36.7              & 76.4        & 20.332  \\
1.5         & 34.7              & 75.7        & 21.154  \\
\bottomrule
\end{tabular}
\label{tab:rp}
\end{table}

\subsubsection{Dynamic Position Shifting}
Positional embeddings are added to the token feature in models like GPT-Neo. During training, this is added per batch of sentences, causing the same sentence to have different offsets in positional embedding in different training batches and during generation.
To improve the extraction of memorized suffixes, we propose to recover the positions used during training by evaluating different shifted positions and selecting the 'best' one. Specifically, for a given prefix $p$, we evaluate different position $\mathcal{C} = {c^i}$, where $c^i$ is a list of consecutive natural numbers, $c^i=\{c^i_1,\cdots\}$, s.t. $|c^i|=|p|$ and calculate the corresponding perplexity values. The position with the lowest perplexity value is then chosen as the position from which to generate the suffix as
\begin{equation}
    c = \mathop{\textrm{argmin}}\limits_{c^i \in \mathcal{C}} \mathcal{P}(p,c^i)\textrm{; }\hspace{0.2cm}\hat{\phi}{(x_i)} = \psi(c_n) + \phi(x_n)\textrm{,} 
\end{equation}
where $\psi(\cdot)$ denotes positional encoding layers, $\phi(\cdot)$ denotes the feature mapping function, $\hat{\phi}$ denotes the feature mapping function consisting positional encoding, and $\mathcal{P}$ computes the perplexity of the prefix. The experimental results are presented in Table~\ref{tab: window_len}. $\mathcal{C}=\{[0,1,\cdots,|p|],[1,2,\cdots,|p|+1],\cdots\}$ is evaluated. The data show that, while using position shifting improves the $\mathcal{M}_{\textrm{H}}$ metric, it may have a negative impact on precision and recall.

\subsection{Look-Ahead}
Table~\ref{tab:fail case} highlights a common issue encountered during the training data extraction process, where only one or two tokens are incorrectly generated or placed in an inappropriate position. To address this problem, we propose a technique that involves looking $\nu$ steps ahead and using the probability of the subsequent tokens to inform the generation of the current token. The goal of look-ahead is to use the posterior distribution to help compute the current token generation probability. We begin by presenting the precise mathematical formulation of the optimal probability and then introduce the implementation, which employs an estimation due to efficiency considerations. The posterior is calculated as
\begin{equation}
\begin{split}
    &f_{\theta}\left(x_{n}|x_{n+1}, x_{<n}\right)\\
=&\frac{f_{\theta}\left(x_{n+1}|x_{n},x_{<n}\right)f_{\theta}\left(x_{n}|x_{<n}\right)}{\sum_{x_{n}'}f_{\theta}\left(x_{n+1}|x_{n}',x_{<n}\right)f_{\theta}\left(x_{n}'|x_{<n}\right)}\textrm{.}
\end{split}
    \label{eq:lookahead_step1}
\end{equation}

\begin{table}[t]
\centering
\small
\caption{Results of $\mathcal{M}_{\textrm{P}}$, $\mathcal{M}_{\textrm{R}}$, and $\mathcal{M}_{\textrm{H}}$ under context window length adjustments. All results are reported on a single trial.}
\vspace{0.1in}
\begin{tabular}{lccc}
\toprule
& $\mathcal{M}_{\textrm{P}}$ (\%)$\left(\uparrow\right)$& $\mathcal{M}_{\textrm{R}}$ (\%)$\left(\uparrow\right)$& $\mathcal{M}_{\textrm{H}}$ $\left(\downarrow\right)$  \\ \midrule
Baseline              & 19.5     & 65.6        & 26.948  \\ \midrule
\rowcolor{Gray} Context Win $\mathcal{W}_w$ & 47.4     & 77.6        & 16.993  \\ 
Context Win $\mathcal{W}_v $       & 46.7     & 77.5        & 17.164 \\\midrule
Position Shifting       & 16.4     &  39.0       & 21.154 \\
\bottomrule
\end{tabular}
\label{tab: window_len}
\end{table}

\begin{table}[t]
\centering
\caption{Results of $\mathcal{M}_{\textrm{P}}$, $\mathcal{M}_{\textrm{R}}$, and $\mathcal{M}_{\textrm{H}}$ under auto-tuning. All results are reported on 5 trials.}
\vspace{0.1in}
\small
\begin{tabular}{lccc}   
\toprule
\textbf{Strategy} & \makecell[c]{$\mathcal{M}_{\textrm{P}}$ (\%)$\left(\uparrow\right)$}&\makecell[c]{$\mathcal{M}_{\textrm{R}}$ (\%)$\left(\uparrow\right)$} & \makecell[c]{$\mathcal{M}_{\textrm{H}}$$\left(\downarrow\right)$ } \\ \midrule
{Baseline}                                                                             & 37.0      & 76.5        & 19.614  \\
Manual selection  & 48.8     & 76.4        & 16.379  \\
{Auto-tuning}                                                                           & 49.4     & 76.6        &  16.127   \\
    \bottomrule
\end{tabular}\label{tab:autotuning}
\end{table}



More generally, let $\textbf{Track}(x_{\textrm{start}}, x_{\textrm{end}}|x_{\textrm{cond}})$ be the probability product of the track starting from $x_{\textrm{start}}$ and ending at $x_{\textrm{end}}$, conditioned on $x_{\textrm{cond}}$. Then we can write $\nu$-step posterior as
\begin{equation}
      \!\!\!\!  f_{\theta}\left(x_{n}|x_{n+\nu}, x_{<n}\right)=\frac{\textbf{Track}(x_{n}, x_{n+\nu}|x_{<n})}{\sum_{x_{n}'}\textbf{Track}(x_{n}', x_{n+\nu}|x_{<n})}\textrm{,}
\end{equation}
where $\textbf{Track}$ is calculated as,
\begin{equation}
\begin{split}
\textrm{\textbf{Track}}(x_{\textrm{start}},x_{\textrm{end}}|x_{\textrm{cond}})&\\
=\sum_{x_{\textrm{start}+1}'}\sum_{x_{\textrm{start}+2}'}...\sum_{x_{\textrm{end}-1}'}&f_\theta(x_{\textrm{start}}|x_{\textrm{cond}})\\
&f_\theta(x_{\textrm{start}+1}|x_{\textrm{cond}},x_{\textrm{start}})...\\
f_\theta(x_{\textrm{end}}|x_{\textrm{cond}}&,x_{\textrm{start}},x_{\textrm{start}+1}'...,x_{\textrm{end}-1}')\textrm{.}
\end{split}
\end{equation}
Based on the 1-step look-ahead formulation in Eq.~(\ref{eq:lookahead_step1}), we utilize the posterior probability to calculate the confidence of the current token. However, due to the size of the vocabulary list (more than 50,000), we only select $\lambda$ tokens whose probability $f_\theta\left(x_{n}'|x_{<n}\right)$ ranks among the highest, as
\begin{equation}
\begin{split}
    \mathcal{X} = \{x_n'|\mathcal{R}( f_\theta\left(x_{n}'|x_{<n}\right)) \geq \lambda\}\textrm{,}
\end{split}
\label{eq:lookahead_approx}
\end{equation}
where $\mathcal{R}(\cdot)$ is the rank function. The experimental results, as detailed in Table~\ref{tab: window_len}, show that using a look-ahead strategy leads to a significant improvement in precision. Furthermore, an increase in the value of $\lambda$ results in a corresponding improvement in performance. It is important to note, however, that increasing the value of $\lambda$ also increases the computational cost, with a complexity of $\mathcal{O}(\lambda\cdot N)$, where $N$ denotes the length of the generated tokens.




\begin{table}[t]
\centering
\small
\caption{Hyper-parameters selection for auto-tuning. Multiple configurations of final hyperparameters are found to yield equivalent performances, and a representative example is presented.}
\vspace{0.1in}
\begin{tabular}{ccccc}
\toprule
\textbf{Parameters}  & \textbf{Range}  & \textbf{Step}  & \textbf{Initial} & \textbf{Final} \\ \midrule
Top-$k$        & {[}1, 50{]}    & 1        & 10    & 24 \\
Nucleus-$\eta$        & {[}0.1, 1{]}  & 0.01     & 0.6   & 0.8\\
Typical-$\phi$   & {[}0.1, 1{]}   & 0.01       & 0.6   & 0.9 \\
Temperature $T$  & {[}0.1, 5{]}   & 0.1       & 0.3   & 0.58 \\
Repetition Penalty & {[}0.8, 1.3{]} & 0.01 & 1 & 1.04 \\ \bottomrule
\end{tabular}\label{tab:autotune para}
\end{table}

\begin{table}[t]
\centering
\small
\caption{Experimental results of $\mathcal{M}_{\textrm{P}}$, $\mathcal{M}_{\textrm{R}}$, and $\mathcal{M}_{\textrm{H}}$ under look-ahead. All results are reported on a single trial.}
\vspace{0.1in}
\begin{tabular}{llccc}
\toprule
&      & $\mathcal{M}_{\textrm{P}}$(\%)$\left(\uparrow\right)$& $\mathcal{M}_{\textrm{R}}$(\%)$\left(\uparrow\right)$  & $\mathcal{M}_{\textrm{H}}$ $\left(\downarrow\right)$ \\ \midrule
\multicolumn{2}{l}{Baseline}                                                                      & 19.5     & 65.6        & 26.948       \\ \midrule
\multicolumn{1}{l}{\multirow{3}{*}{Look-ahead}} & $\lambda$=10 & 33.1     & 71.6        & 24.262   \\
\multicolumn{1}{l}{}                                                                       & $\lambda$=20 & 35.5     & 72.6        & 22.157    \\
\multicolumn{1}{l}{}                                                                       & $\lambda$=30 & 36.7     & 73.0        & 21.333    \\
    \bottomrule
\end{tabular} \label{tab:lookahead}
\end{table}

\subsection{Hyperparameter Optimization}
The aforementioned tricks in Section~\ref{sect:sampling} involve various hyperparameters, and simply using the best parameters is usually suboptimal. Manually searching for the best hyperparameters, also known as 'babysitting,' can be time-consuming. We use a versatile architecture auto-tuning method~\citep{akiba2019optuna}, which incorporates efficient search and pruning strategies, to determine the optimized hyperparameters following advanced frameworks~\citep{snoek2012practical,koch2018autotune,akiba2019optuna}. As the search algorithm, we use covariance matrix adaptation evolutionary strategies (CMA-ES)~\citep{hansen2003reducing}. The search objective in our experiment is set to $\mathcal{M}_{\textrm{P}}$, and the parameters that are searched over include top-$k$, nucleus-$\eta$, typical-$\phi$, temperature $T$, and repetition penalty $r$.

Table~\ref{tab:autotune para} contains the detailed search parameter settings. We also provide the baseline parameters as initial values to the search algorithm to speed up convergence. The number of search rounds is limited to 1,000. The experimental results are shown in Table~\ref{tab:autotuning}. Simply using the best parameters outlined in Section.~\ref{sect:sampling} with $k$=5, $\eta$=0.6, $\phi$=0.6, $T$=0.4, $r$=1 yields a precision of 48.8\%. Implementing auto-tuning results in a 37\% improvement over baseline, and auto-tuning performs slightly better than the hypermeter manual section.


\vspace{-0.cm}
\section{Improved Suffix Ranking} \label{sect: ranking}
\vspace{-0.0cm}
Following the generation of multiple suffixes, a ranking process is carried out in which less likely suffixes are eliminated using perplexity $\mathcal{P}$ as a metric. However, our statistical analysis in Figure~\ref{fig:sen_hist_original} reveals that the ground-truth sentences do not consistently have the lowest perplexity values. Thus, we propose additional ranking metrics to address this disparity.

\begin{table}[t]
\centering
\caption{Experimental results of precision, recall, and Hamming distance under different ranking methods. All results are reported on a single trial.}
\vspace{0.05in}
\small
\begin{tabular}{llll}
\toprule
   \textbf{Method}             &\makecell[c]{$\mathcal{M}_{\textrm{P}}$(\%)$\left(\uparrow\right)$}    & \makecell[c]{$\mathcal{M}_{\textrm{R}}$(\%)$\left(\uparrow\right)$}  & \makecell[c]{ $\mathcal{M}_{\textrm{H}}$$\left(\downarrow\right)$}      \\ \midrule
 Baseline           & 37.0  & 76.5 & 19.614 \\ \midrule
Perplexity\,$\div$\,Zlib    & 37.1 & 76.0 & 20.191 \\
 Perplexity\,$\times$\,Zlib    & 37.1 & 76.3 & 20.368 \\
 Cumprod           & 36.6 & 39.8  & 20.127 \\\midrule
High confidence    & 40.6 & 77.3 & 19.518    \\
 Surprised patterns & 37.1 & 75.5 & 20.072     \\
    \bottomrule
\end{tabular} \label{tab:membership}
\end{table}

\begin{table}[t]
\centering
\vspace{-0.4cm}
\small
\caption{Experimental results of precision, recall, and Hamming distance for compatibility analysis. All results are reported on a single trial.}
\vspace{0.05in}\label{tab:gptneo-27}
\resizebox{1.0\linewidth}{!}{
\begin{tabular}{llll} 
\toprule
& $\mathcal{M}_{\textrm{P}}$(\%) $\uparrow$ & $\mathcal{M}_{\textrm{R}}$(\%) $\uparrow$ & $\mathcal{M}_{\textrm{H}}$ $\downarrow$ \\ \midrule
Baseline(GPT-Neo 1.3B) & 19.5 & 65.6 & 26.948 \\
Context win\,+\,High confidence(GPT-Neo 1.3B)   & 46.8  & 77.5  & 17.144  \\
Baseline(GPT-Neo 2.7B) & 33.5 & 76.6 & 20.921 \\
Context win\,+\,High confidence(GPT-Neo 2.7B)   & 54.8  & 81.5  & 13.621  \\
\bottomrule 
\end{tabular}
}
\end{table}

\subsection{Sentence-Level Criteria}
\textbf{Zlib.} The entropy of a text, as determined by the Zlib compression algorithm~\citep{gailly2004zlib} using the number of bits, is a quantitative indicator of the sequence's information content. We use the ratio of the perplexity of a given sentence as computed by the GPT-Neo model, and the Zlib entropy of the same sentence as a metric for membership inference, as outlined in the work of \citet{carlini2021extracting}. Besides, we investigate the potential utility of producting perplexity and Zlib entropy, as both metrics tend to decrease when the model shows a high degree of confidence in its predictions. Both metrics produce only a marginal improvement in the overall performance of the membership inference task, as shown in Table~\ref{tab:membership}.

\textbf{Cumprod.} We consider alternative metrics such as the cumprod, which depicts the tandem probability of a generation sentence as $\mathcal{L}_c = (\prod_{n=0}^{N} \log p(x_n|x_0,..,x_{n-1}))^{-N}$. However, the effect of tandem probability on precision is marginal and has a negative effect on $\mathcal{M}_{\textrm{R}}$.

\subsection{Token-Level Criteria}

\textbf{Reward on high confidence.} The presence of a high degree of confidence in memorized data is one of the defining features of the phenomenon known as the'memorization effect'~\citep{goodfellow2016deep,zhang2021understanding,pmlr-v70-arpit17a}.  We propose implementing a strategy that rewards suffixes with high-confidence tokens based on this insight. If the sentence contains confident tokens, the possibility of the generated token is higher than a threshold, and the difference between the generated token and other token is higher than a threshold, we rank it higher. Specifically, for the token $x_n$ in a generated suffix, if the probability of is higher than a threshold $0.9$, then we subtract a given number $0.1$ from the score of suffix $s_i$ (the original score for $s_i$ is its perplexity). This trick makes a noticeable improvement as seen in Table~\ref{tab:membership}.

\begin{table}[t]
\centering
\small
\caption{Experimental results of precision, recall, and Hamming distance for compatibility analysis. All results are reported on a single trial.}
\vspace{0.05in}\label{tab:compatible}
\resizebox{1.0\linewidth}{!}{
\begin{tabular}{llll} 
\toprule
& $\mathcal{M}_{\textrm{P}}$(\%) $\uparrow$ & $\mathcal{M}_{\textrm{R}}$(\%) $\uparrow$ & $\mathcal{M}_{\textrm{H}}$ $\downarrow$ \\ \midrule
Context win\,+\,Beams=2 & 46.7  & 77.5  & 17.154  \\
Auto-tuning\,+\,Beams=2 & 46.4 & 76.2 & 17.331 \\
Context win\,+\,Auto-tuning   & 46.5  &  77.5 & 17.370  \\
Context win\,+\,High confidence   & 46.8  & 77.5  & 17.144  \\
\bottomrule 
\end{tabular}
}
\end{table}

\textbf{Encouraging surprised patterns.} According to recent studies~\citep{holtzman2019curious}, human text generation frequently exhibits a pattern in which tokens with high perplexity are intermittently included, as opposed to consistently selecting tokens with low perplexity. To address this problem, we propose a simple solution that encourages the presence of surprised tokens (high perplexity tokens) by calculating the perplexity of a generated prompt based on only the majority tokens:
\begin{equation} \label{eq:sp}
\setlength\abovedisplayskip{3pt}
\setlength\belowdisplayskip{3pt}
\scalemath{0.95}{
    \begin{aligned}
    \mathcal{L}_s = \frac{1}{|\hat{\mathcal{X}}|}&\sum_{x_n \in \hat{\mathcal{X}}} \log p(x_n|x_{[0:n-1]})\textrm{,} \\
    \hat{\mathcal{X}} = \{x_j| \mu-3\sigma < &\log p(x_n|x_{[0:n-1])}) < \mu+3\sigma\}\textrm{,}
        \end{aligned}}
\end{equation} 
where $\mu$ and $\sigma$ denotes the mean and standard of the $p(x_n|x_{[0:n-1]})$ in a batch. The inclusion of surprised tokens in a generation does not have a negative impact on overall sentence perplexity when using this method, thereby increasing the likelihood of their selection during membership inference. As demonstrated in Table~\ref{tab:membership}, the improvement is relatively marginal.


\subsection{Compatibility Analysis}
It has been discovered that accumulating the aforementioned tricks does not always result in a proportionate increase in performance. We investigate the interactive effects of combining various techniques in a targeted and efficient manner, in order to establish a viable baseline method. The results are shown in Table~\ref{tab:compatible}. 
The application of multiple of the aforementioned techniques yields no significant improvement in performance. Even when using a beam width of 2 and maintaining the same settings as in the previous experiments, the results do not show a significant improvement. The empirical findings presented in this section suggest that to achieve optimal results, a more deliberate and strategic combination of various methods is required.

We also evaluate the methods on GPT-Neo 2.7B in Table~\ref{tab:gptneo-27} and draw the conclusion that a larger pre-trained model yield better results.

\section{Conclusion and Future Work}
We investigate a dozen techniques for extracting training data from LMs. These techniques make minor changes to the generation strategies and ranking strategies. Our empirical findings show that several of these tricks improve significantly, while interactions between different tricks are more subtle than expected. Besides, the empirical results show that commonly used versatile methods for general text generation are not always effective for extraction tasks. 
In future works, explortion in developing compatible techniques in data extraction is preferred. Techniques in both suffix generation and ranking suffix may be combined and explored in an end-to-end fashion. In addition, further investigation to gain a deeper understanding of the underlying mechanisms that contribute to the superior performance of certain techniques in this paper are encouraged.

\bibliography{lm}
\bibliographystyle{icml2018}

\newpage
\appendix
\onecolumn
\section{Appendix}

\subsection{Guidelines} \label{sect: principle}
Recently, there has been a growing emphasis on the importance of extracting training data for language models. Through our examination of various techniques in this paper, we have discovered that certain tricks do not yield as favorable results as compared to the baseline approach, or may entail substantial computational expense. Therefore, we propose the following guidelines for the development of data extraction methods for language models. 
\begin{enumerate}
    \item Usability: In order for a data extraction method to be considered practical, it is expected have a low threshold for adoption and be easy to use. For example, not requiring additional datasets, having a straightforward implementation without the need for complex modules, and being easily adaptable for various applications.
    \item Effectiveness: The data extraction methods are expected to demonstrate a good performance on the given datasets.
    \item Efficiency: It is essential that the methods are computationally efficient, utilizing limited computation cost. Additionally, it is desirable for the methods to incorporate small yet effective modules. Extra LMs improve the performances, but we expect the method to replace them with small modules without a noticeable performance trade-off.
    \item Perniciousness. As an adversarial task to data protection, we expect data extraction methods from language models to be pernicious since a pernicious strong extraction attack can better evaluate the risk of information leakage accurately, equivalent to a higher lowerbound for potential leakage.
\end{enumerate}
We would like to emphasize that the guidelines aim to enhance the success rate of the training data extraction methods and thus provide a more accurate evaluation of the privacy information leakage of LMs, instead of posing pernicious attack methods.

\subsection{Typical Failure Cases of Training Data Extraction}
In this section, we observe the classical instances of failures that occur in extracting training data and classify them into two categories, as outlined in Table~\ref{tab:fail case}. The first category of failure  pertains to instances where the generated suffix is incorrect from the tokens in the middle portion, resulting in subsequent tokens being generated inaccurately. The second category of failure pertains to instances where only a limited number of tokens in the suffix do not correspond with the ground truth and the latter tokens are correctly generated. 

The method of look-ahead is motivated by the second category of failure encountered during training data extraction. In addition, it is evident from the examples of both successful and failure cases of data extraction that, both the prefix and the suffix are not complete sentences. For instance, the prefix can begin with a punctuation mark `.',  or a broken sentence like 'WARRANTY OF ANY KIND'. It is due to the prevalent techniques of truncating and padding training language materials. In light of this phenomenon, this paper presents the methods of dynamic context window and dynamic position shifting.

In contrast to failure cases, Table~\ref{tab: successful eg} presents a selection of successful examples of extraction. These examples may serve to provide a tangible understanding of the task of training data extraction.

\subsection{Histogram without Interploration}
\subsubsection{Sentence perplexity histogram}
To investigate the distribution of sentence perplexity of generated suffixes and ground truth suffixes, we use histograms to visualize their discrepancies. 
We compute the sentence perplexity of both ground truth and multiple generated suffixes, and then calculate the rank of ground truth perplexity and draw a histogram based on this in Figure~\ref{fig:sen_hist_original}, where the x-axis is calculated as 
\begin{equation}
    \mathcal{S}(\mathcal{P}_{gt},\{\mathcal{P}_{s_i}\})\textrm{,}
\end{equation}
where $\mathcal{S}(a,b)$ returns the rank of a in a set b, $\mathcal{P}_{gt}$ is the perplexity of the ground truth suffix of a given prefix $p$,  $s_i$ denotes the i-th generated suffix of a given prefix $p$. 

\subsubsection{Token logits histogram}
\begin{figure}[ht]
\centering
    \begin{minipage}[ht]{0.45\textwidth}
        \centering
        \includegraphics[width=1\textwidth]{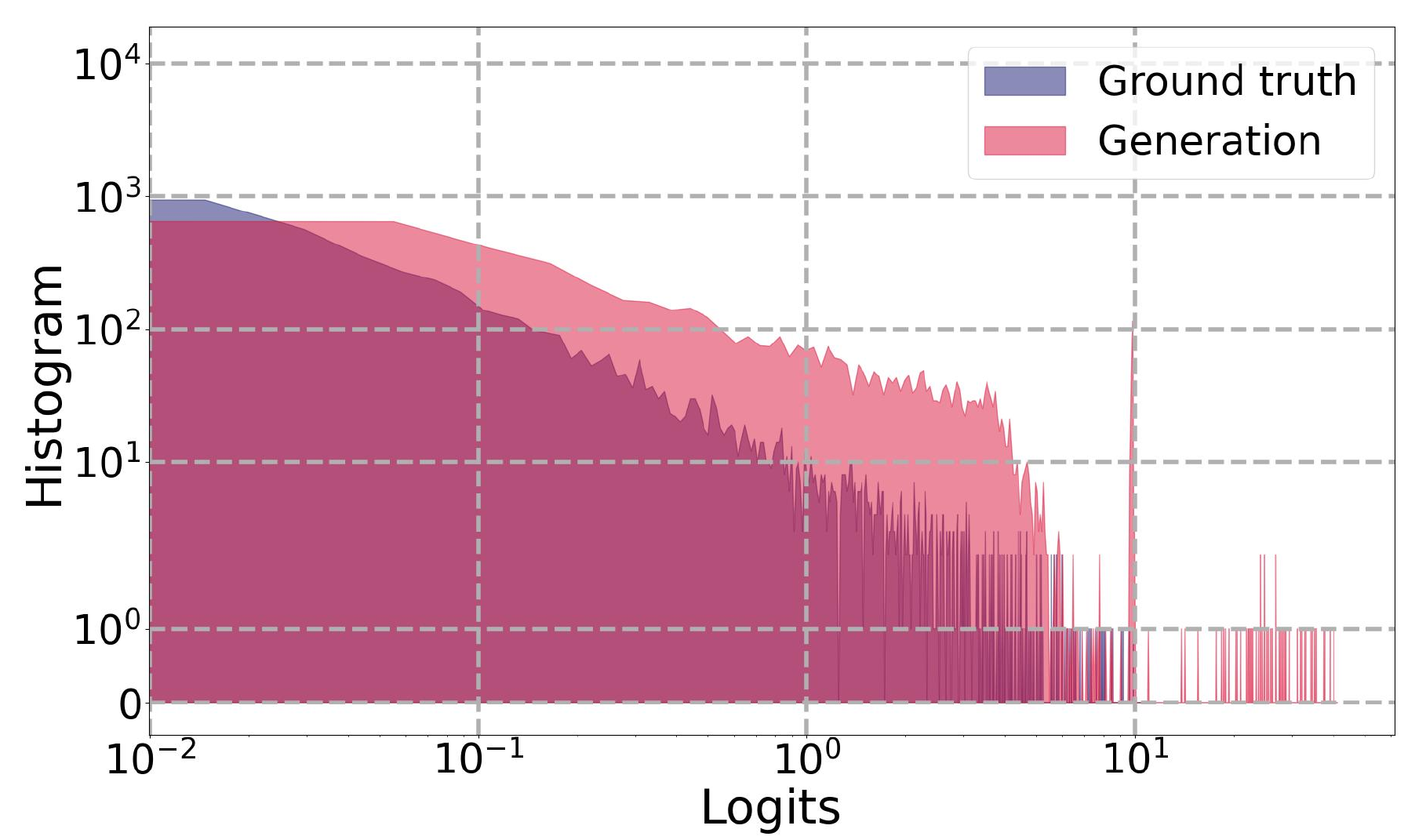}
        \caption{Histogram of token logits. The histogram depicts the distribution of logit values obtained from 1,000 suffixes, each containing 50 tokens.}
        
        \label{fig:token_hist}
    \end{minipage}
    \begin{minipage}[ht]{0.45\textwidth}
        \centering
        \includegraphics[width=1\textwidth]{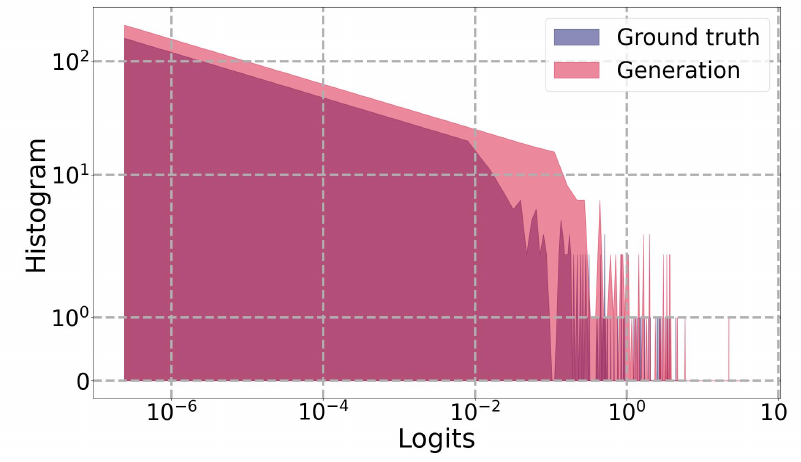}
        \caption{Logits values of 0-th (the first) generated token. The histogram depicts the distribution of logit values obtained from the 0-th token of 1,000 suffixes.}
        \label{fig:token_hist_0}
        
    \end{minipage}

            \begin{minipage}[ht]{0.45\textwidth}
        \centering
        \includegraphics[width=1\textwidth]{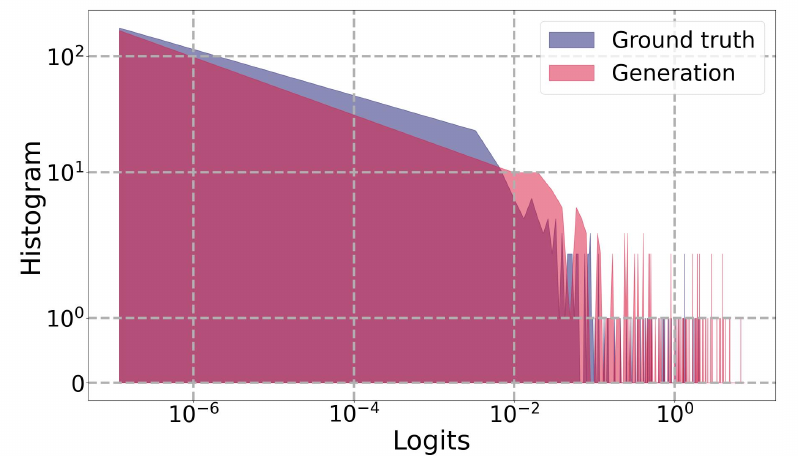}
        \caption{Logits values of 24-th (the middle) generated token. The histogram depicts the distribution of logit values obtained from the 24-th token of 1,000 suffixes.}
        \vspace{-1em}
        \label{fig:token_hist_24}
    \end{minipage}
    \begin{minipage}[ht]{0.45\textwidth}
        \centering
        \includegraphics[width=1\textwidth]{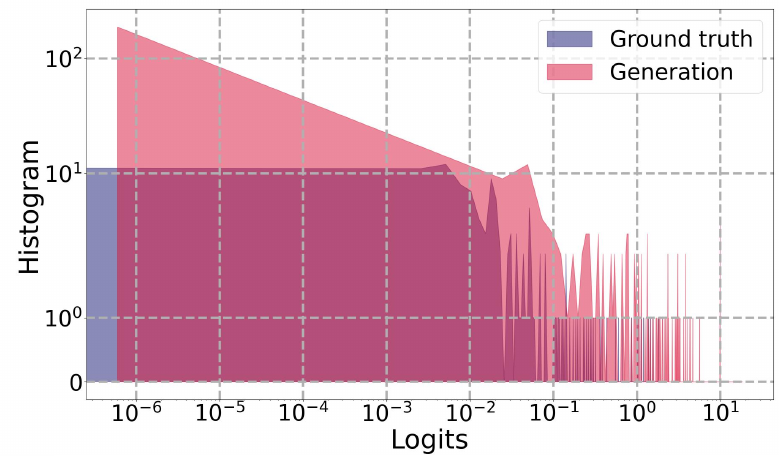}
        \caption{Logits values of 49-th (the last) generated token. The histogram depicts the distribution of logit values obtained from the 49-th token of 1,000 suffixes.}
        \label{fig:token_hist_49}
        \vspace{-1em}
    \end{minipage}
\end{figure}

To investigate the distribution of logits values between generated tokens and ground truth tokens, we use histograms to visualize their discrepancies. 
A histogram of token logits for both ground truth suffixes and generated suffixes is presented in Figure~\ref{fig:token_hist}, wherein a clear discrepancy in the logits distribution can be observed. Furthermore, in order to gain insight into the distribution of logits values across different positions, histograms of logits values for various positions are also provided in Figure~\ref{fig:token_hist_0},~\ref{fig:token_hist_24}, and~\ref{fig:token_hist_49}.

\begin{table}[ht]
\centering
\setlength{\tabcolsep}{3pt}
\caption{Training data extraction successful cases. The generated suffix equals the ground truth suffix in both length and tokens.}\label{tab: successful eg}
\vspace{0.1in}
\begin{tabular}{cl}
\toprule
 Prefix     & \makecell[l]{/*
\\ * Copyright 2003-2006 Sun Microsystems, Inc.  All Rights Reserved.
\\ * DO NOT ALTER OR REMOVE COPYRIGHT NOTICES OR THIS FILE HEADER.
\\ *
\\ * This code is free software; }\\ \midrule
 \begin{tabular}[c]{@{}c@{}}Generated \\ Suffix\end{tabular}   & \makecell[l]{MERCHANTABILITY or
\\ * FITNESS FOR A PARTICULAR PURPOSE.  See the GNU General Public License
\\ * version 2 for more details (a copy is included in the LICENSE file that
\\ * accompanied this code).}  \\ \midrule

 Prefix     & \makecell[l]{
Permission is hereby granted, free of charge, to any person obtaining a copy of this software and associated \\documentation files (the \&quot;Software\&quot;), to deal in the Software without restriction, including \\without limitation the rights to}\\ \midrule
 \begin{tabular}[c]{@{}c@{}}Generated \\ Suffix\end{tabular}   & \makecell[l]{ INCLUDING BUT NOT LIMITED TO THE WARRANTIES OF MERCHANTABILITY, FITNESS FOR A \\PARTICULAR PURPOSE AND NONINFRINGEMENT. IN NO EVENT SHALL THE AUTHORS OR \\COPYRIGHT HOLDERS BE LIABLE FOR
}  \\ 
    \bottomrule
\end{tabular}
\end{table}

\subsection{Takeaways}
we conclude with some takeaways as follows,
\begin{enumerate}
    \item In the two stages (suffix generation and suffix ranking) of training data extraction, suffix generation influences the extraction precision more than suffix ranking.
    \item Dynamic context window is a significantly more useful trick than others.
    \item Sampling strategies are highly sensitive to hyperparameters.
    \item We recommend using fewer than or equal to two beams, as the number of beams significantly increases memory requirements but brings degressive precision improvements.
    \item The look-ahead mechanisms augments the precision of the presently generated token, and the degree of enhancement is contingent upon the number of future tokens taken into account.
\end{enumerate}



\subsection{Negative Societal Impacts}

In this study, various techniques for extracting training data are investigated. It is imperative to note that these methods may be utilized for nefarious purposes. The focus of this paper is on a specific type of training data extraction, namely the target training data extraction, which requires a prefix in the training dataset. If an attacker has access to the training data of a language model or the ability to replicate similar sentences to the training data, it is possible for them to recover the sentences of the training data, as demonstrated in Table~\ref{tab: successful eg}.

We mitigate the negative ethics impacts in this work, where we focus on a specific LM GPT-Neo, which uses Pile as its training dataset. Note that Pile is already a public dataset obtained from mainly academic or professional sources instead of private sources. 

The primary intended users of training data extraction include institutions and organizations that aim to investigate the privacy and security characteristics of language models. However, it is crucial to acknowledge that these methods may also be used by malicious attackers. In the short term, some LMs present a risk of leaking information. And in the long term, we appeal to the language community to mitigate the risk and develop secure language models which exhibit resistance to attacks including membership inference attacks, training data extraction attacks, etc.

\begin{table}[htbp]
\centering
\setlength{\tabcolsep}{3pt}
\caption{Training data extraction examples. {\color{red}{Red}} tokens indicate the mismatch tokens between the generated suffix and ground truth suffix.}\label{tab:fail case}
\vspace{0.1in}
\begin{tabular}{cl}
\toprule
 Prefix     & \makecell[l]{.\\
//\\
// Copyright (c) 2008-2011 Texas Instruments Incorporated.  All rights reserved.\\
// Software License Agreement\\
// \\
// Texas Instruments (TI) is supplying this software for use solely and\\
// exclusively on TI }\\ \midrule
 \begin{tabular}[c]{@{}c@{}}Generated \\ Suffix\end{tabular}   & \makecell[l]{ NOT LIMITED TO, IMPLIED WARRANTIES OF MERCHANTABILITY AND FITNESS FOR
\\// A PARTICULAR PURPOSE APPLY TO THIS SOFTWARE.
\\// \color{red}{Redistribution and use insource and binary forms, with}}  \\ \midrule 
 \begin{tabular}[c]{@{}c@{}}Ground Truth\\ Suffix\end{tabular} & \makecell[l]{ NOT LIMITED TO, IMPLIED WARRANTIES OF MERCHANTABILITY AND FITNESS FOR
\\// A PARTICULAR PURPOSE APPLY TO THIS SOFTWARE. \color{red}{TI SHALL NOT, UNDER ANY}
\\// \color{red}{CIRCUMSTANCES, BE LIA}}\\\midrule

Prefix                                                        & \makecell[l]{\\.
\\No comments:
\\Friends
\\Dis-complainer
\\The Great Change is published whenever the spirit moves me. Writings on this site are purely the opinion\\ of Albert Bates and are subject to a Creative Commons Attribution Non-}  \\ \midrule 
 \begin{tabular}[c]{@{}c@{}}Generated \\ Suffix\end{tabular}   & \makecell[l]{ (NC): You may not use this work for commercial purposes. \color{red}{Commercial use is not permitted without the} \\ \color{red}{express written consent of the copyright holder.}}   \\ \midrule
 \begin{tabular}[c]{@{}c@{}}Ground Truth\\ Suffix\end{tabular} & \makecell[l]{ (NC): You may not use this work for commercial purposes. \color{red}{Share Alike (SA): If you alter, transform, or }\\\color{red}{build upon this work, you may distribute the resulting work only under the same or similar license to this}\\\color{red}{ one. Nothing in}} \\ \midrule

Prefix                                                        & \makecell[l]{
\# Permission is hereby granted, free of charge, to any person
\\\# obtaining a copy of this software and associated  documentation
\\\# files (the "Software"), to deal in the Software without
\\\# restriction, }  \\ \midrule 
 \begin{tabular}[c]{@{}c@{}}Generated \\ Suffix\end{tabular}   & \makecell[l]{WARRANTY OF ANY KIND,
\\\# EXPRESS OR IMPLIED, INCLUDING BUT NOT LIMITED TO THE WARRANTIES \color{red}{OF} \\\# MERCHANTABILITY, FITNESS FOR A PARTICULAR PURPOSE AND
\\\# NONINFRINGEMENT   }   \\ \midrule
 \begin{tabular}[c]{@{}c@{}}Ground Truth\\ Suffix\end{tabular} & \makecell[l]{WARRANTY OF ANY KIND,
\\\# EXPRESS OR IMPLIED, INCLUDING BUT NOT LIMITED TO THE WARRANTIES
\\\# {\color{red}{OF}} MERCHANTABILITY, FITNESS FOR A PARTICULAR PURPOSE AND
\\\# NONINFRINGEMENT} \\ \midrule

Prefix                                                        & \makecell[l]{            ////
\\////     THIS SOFTWARE IS PROVIDED ``AS IS'' AND WITHOUT ANY     ////
\\//// EXPRESS OR IMPLIED WARRANTIES, INCLUDING}  \\ \midrule 
 \begin{tabular}[c]{@{}c@{}}Generated \\ Suffix\end{tabular}   & \makecell[l]{ DAMAGES   ////
\\//// (INCLUDING, BUT NOT LIMITED TO, PROCUREMENT OF SUBSTITUTE {\color{red}{GOODS}}  ////
\\//// OR SERVICES; LOSS OF USE, DATA, OR PROFITS; {\color{red}{OR BUS}} }   \\ \midrule
 \begin{tabular}[c]{@{}c@{}}Ground Truth\\ Suffix\end{tabular} & \makecell[l]{ DAMAGES    ////
\\//// (INCLUDING, BUT NOT LIMITED TO, PROCUREMENT OF SUBSTITUTE   ////
\\//// {\color{red}{GOODS}} OR SERVICES; LOSS OF USE, DATA, OR PROFITS;} \\ 
    \bottomrule
\end{tabular}
\end{table}

\end{document}